\documentclass[sigconf, nonacm, screen, review=false]{acmart}

\usepackage{multirow}   
\usepackage{enumitem}
\usepackage{framed}
\usepackage{orcidlink}
\usepackage{acmart-taps}
\usepackage{xcolor}

\usepackage[fixed]{fontawesome5}

\usepackage{hyperref}
\usepackage[nameinlink]{cleveref}

\usepackage{fancyhdr}
\usepackage{graphicx}
\usepackage{booktabs}

\usepackage{subcaption}

\definecolor{clrGray}{RGB}{209,207,199}
\definecolor{clrGrayDark}{RGB}{95,94,90}
\definecolor{clrPurple}{RGB}{175,169,236}
\definecolor{clrPurpleDark}{RGB}{83,74,183}
\definecolor{clrTeal}{RGB}{93,202,165}
\definecolor{clrTealDark}{RGB}{15,110,86}
\definecolor{clrBlue}{RGB}{133,183,235}
\definecolor{clrBlueDark}{RGB}{24,95,165}
\definecolor{clrAmber}{RGB}{239,159,39}
\definecolor{clrAmberDark}{RGB}{133,79,11}
\definecolor{clrPhase}{RGB}{240,238,232}
\colorlet{shadecolor}{clrPhase}

\definecolor{authority}{HTML}{6A4C93}
\definecolor{care}{HTML}{00B4A6}
\definecolor{fairness}{HTML}{3498DB}
\definecolor{loyalty}{HTML}{E74C3C}
\definecolor{sanctity}{HTML}{F39C12}

\AtBeginDocument{%
  }

\AtBeginDocument{
  \setlength{\textfloatsep}{8pt plus 2pt minus 2pt}
}

\copyrightyear{2026}
\acmYear{2026}
\setcopyright{cc}
\setcctype{by}
\acmConference[GoodIT '26]{International Conference on Information Technology for Social Good}{September 02--04, 2026}{Pisa, Italy}
\acmBooktitle{International Conference on Information Technology for Social Good (GoodIT '26), September 02--04, 2026, Pisa, Italy}
\acmDOI{10.1145/3794786.3830756}
\acmISBN{979-8-4007-2483-1/2026/09}

\begin{document}

\title{Moral Semantics Survive Machine Translation}
\subtitle{Cross-Lingual Evidence from Moral Foundations Corpora}

\author{Maciej Sk\'{o}rski \orcidlink{0000-0003-2997-7539}}
\orcid{0000-0003-2997-7539}
\email{maciej.skorski@gmail.com}
\affiliation{%
  \institution{University of Luxembourg}
  \department{Faculty of Science, Technology and Medicine}
  \city{Luxembourg}
  \country{Luxembourg}
}
\affiliation{%
  \institution{University of Warsaw}
  \department{Institute of Informatics}
  \city{Warsaw}
  \country{Poland}
}

\begin{abstract}
Moral language is subtle and culturally variable, making it
difficult to translate faithfully across languages. Idiomatic
expressions, slang, and cultural references introduce hard-to-avoid translation
artifacts. Yet automated moral values
classification depends on language-specific annotated corpora that
exist almost exclusively in English. 

We investigate whether LLM-based translation can bridge this
gap, taking Polish as a test case. Using $\sim$50k morally-annotated
social media posts from a diverse range of topics, we apply a
principled four-method validation pipeline: LaBSE cross-lingual
embedding similarity, Centered Kernel Alignment (CKA), LLM-as-judge
evaluation, and deep learning classifier parity tests. We show that
despite shortcomings in handling slang, vulgarity, and
culturally-loaded expressions, direct translation preserves subtle
moral cues well enough to be harvested by cross-lingual machine
learning — with a mean cosine similarity of 0.89 and classification accuracy gaps of
 0.01--0.02 AUROC across foundations.

These results demonstrate that machine translation is a practical
and cost-effective path to moral values research in languages
currently under-resourced in this domain. We demonstrate this for
Polish as a representative Slavic language, with expected
generalization to related languages.
\end{abstract}



\maketitle

\section{Introduction}

Moral language is subtle. Irony inverts it. Cultural idiom obscures
it. Register shifts dilute it. Studying it at scale, across languages
and cultures, requires a principled framework. Moral Foundations
Theory (MFT)~\cite{haidt2004intuitive,graham2013moral} provides
exactly that: a cross-cultural taxonomy of five moral dimensions ---
\textit{care/harm}, \textit{fairness/cheating},
\textit{loyalty/betrayal}, \textit{authority/subversion}, and
\textit{sanctity/degradation} --- grounded in decades of
cross-cultural moral psychology research.

\aptLtoX{
\begin{table}[h]
\centering
\caption{Examples of moral foundations in social media posts~\citep{skorski-landowska-2025-beyond}.
Key moral cues highlighted in foundation colour.}
\label{tab:mft-examples}
\resizebox{0.99\columnwidth}{!}{
\begin{tabular}{p{0.65\columnwidth} p{0.30\columnwidth}}
\toprule
\textbf{Text Example} & \textbf{Moral foundation} \\
\midrule
``My \textcolor{care}{heart breaks} seeing
\textcolor{care}{children separated} from families at the
border'' &
\begin{imageonly}\textcolor{care}{\textbf{\faHandsHelping\ Care}}\end{imageonly} \\[4pt]

``Everyone deserves \textcolor{fairness}{equal access} to healthcare
\textcolor{fairness}{regardless} of income'' &
\begin{imageonly}\textcolor{fairness}{\textbf{\faBalanceScale\ Fairness}}\end{imageonly} \\[4pt]

``\textcolor{authority}{Respect} your elders and follow
\textcolor{authority}{traditional} values that built
this nation'' &
\begin{imageonly}\textcolor{authority}{\textbf{\faCrown\ Authority}}\end{imageonly} \\[4pt]

``\textcolor{loyalty}{Stand with our troops} --- they
sacrifice everything for our freedom'' &
\begin{imageonly}\textcolor{loyalty}{\textbf{\faShield*\ Loyalty}}\end{imageonly} \\[4pt]

``\textcolor{sanctity}{Marriage} is \textcolor{sanctity}{sacred} and should be
protected from secular corruption'' &
\begin{imageonly}\textcolor{sanctity}{\textbf{\faDove\ Sanctity}}\end{imageonly}  \\
\bottomrule
\end{tabular}
}
\end{table}
}
{\begin{table}[h]
\centering
\caption{Examples of moral foundations in social media posts~\citep{skorski-landowska-2025-beyond}.
Key moral cues highlighted in foundation colour.}
\label{tab:mft-examples}
\resizebox{0.99\columnwidth}{!}{
\begin{tabular}{p{0.65\columnwidth} p{0.30\columnwidth}}
\toprule
\textbf{Text Example} & \textbf{Moral foundation} \\
\midrule
``My \textcolor{care}{heart breaks} seeing
\textcolor{care}{children separated} from families at the
border'' &
\textcolor{care}{\textbf{\faHandsHelping\ Care}} \\[4pt]

``Everyone deserves \textcolor{fairness}{equal access} to healthcare
\textcolor{fairness}{regardless} of income'' &
\textcolor{fairness}{\textbf{\faBalanceScale\ Fairness}} \\[4pt]

``\textcolor{authority}{Respect} your elders and follow
\textcolor{authority}{traditional} values that built
this nation'' &
\textcolor{authority}{\textbf{\faCrown\ Authority}} \\[4pt]

``\textcolor{loyalty}{Stand with our troops} --- they
sacrifice everything for our freedom'' &
\textcolor{loyalty}{\textbf{\faShield*\ Loyalty}} \\[4pt]

``\textcolor{sanctity}{Marriage} is \textcolor{sanctity}{sacred} and should be
protected from secular corruption'' &
\textcolor{sanctity}{\textbf{\faDove\ Sanctity}}  \\
\bottomrule
\end{tabular}
}
\end{table}
}

While these foundations are universal, cultures differ markedly in
their sensitivity to each dimension and in how they express it in
language~\cite{graham2013moral}. Automated methods have made it possible to measure these
cultural-linguistic sensitivities at scale. Lexicon-based
approaches~\cite{araque_moralstrength_2020,hoppExtendedMoralFoundations2021} and, more recently, fine-tuned language
models~\cite{nguyenMeasuringMoralDimensions2024a,zangari-etal-2025-me2,
preniqiMoralBERTFineTunedLanguage2024,skorski-landowska-2025-beyond} have been applied
to political discourse~\cite{royAnalysisNuancedStances2021},
social media analysis~\cite{hooverMoralFoundationsTwitter2020}, and moral dilemmas~\cite{nguyen2022mapping} --- building on
foundational cross-cultural findings in moral
psychology~\cite{graham2009liberals,feinberg2013moral}.

Yet these methods require language-specific annotated corpora for
training, and such resources remain almost
exclusively in
English~\cite{tragerMoralFoundationsReddit2022,
hooverMoralFoundationsTwitter2020}. No morally-annotated corpus has
been released for any Slavic language so far --- leaving the moral discourse
of hundreds of millions of speakers beyond the reach of automated
MFT analysis.

Machine translation (MT) offers a natural shortcut: translate existing English corpora and extend MFT tools to new languages at low cost. Automated translation is generally found to preserve meaning, though it distorts tone and style~\cite{tirkkonenCondit2002,wein-schneider-2024-lost}; it can also inject systematic bias into cross-lingual text analysis~\cite{nicholas2023lost}, particularly visible in affective signals \cite{plazadelarco2024angry}. But moral language hits precisely that weak spot --- laden with irony, emotion, cultural idiom, and register sensitivity, it resists the literal mappings that MT systems rely on. This raises a pointed question for moral NLP:

\begin{snugshade}
\small\itshape
Does LLM translation from English (EN) to Polish (PL) preserve  moral-semantics of MFT-annotated texts, despite the subtlety of moral
language?
\end{snugshade}

This study gives a positive answer, contributing:

\begin{itemize}[leftmargin=1em]
  \item \textbf{Principled validation framework.} A reproducible
        multi-method pipeline combining LLM-as-judge evaluation,
        embedding-based similarity (LaBSE, CKA), and deep learning
        classifier parity tests --- applicable to any source language,
        target language, and annotation schema.
  \item \textbf{Large translated corpus.} A validated EN$\to$PL
        translation of $\sim$50k morally-annotated social media posts
        spanning a diverse range of topics and platforms, produced
        using Claude Sonnet at a cost of approximately 200~USD ---
        demonstrating accessibility.
  \item \textbf{Evidence that it works --- and generalizes.}
        Translation is of good quality but not perfect (mean cosine
        0.89), yet classification remains near-parity with English
        originals across five MFT foundations --- typical AUC gaps of
        0.01--0.02 for tuned classification heads and similarly close under full tuning. As one of the most
        morphologically complex Slavic languages~\cite{kann2017oneshot},
        Polish is a demanding test case; success here suggests
        generalization across the broader Slavic family.
\end{itemize}

\section{Background and Related Work}

\subsection{Moral Foundations Basics}

According to MFT, there is a set of universal moral intuitions (foundations), rooted in evolutionary adaptive challenges~\cite{haidt2012righteous}.

Two of them, care and fairness, emphasize, respectively, human welfare and social justice; they are jointly referred to as individualizing foundations as they are focused on protecting the individual.

The three other, on the other hand, are concerned about protecting groups (societal institutions, families, and tribes); these are loyalty, authority, and sanctity, called binding foundations.
Binding foundations emphasize group-binding loyalty and patriotism, respect for authority and tradition, and (spiritual) purity and religion.

Although people share the same cognitive moral learning modules, life experiences and social contexts make them prioritize these differently, resulting in differently guided moral judgments~\cite{graham_moral_2013}.
The care and fairness foundations are believed to be the most universal ones; endorsement of these foundations is high among most people \cite{haidt2012righteous}.
However, high preference for binding moral foundations is seen among collectivistic (e.g. Eastern) cultures \cite{atariMoralityWEIRDHow2023}, religious people \cite{graham_beyond_2010}, and ideological conservatives \cite{haidtWhenMoralityOpposes2007,graham_liberals_2009,kivikangas_moral_2021}, evidence of how moral foundations shape norms and values.
Due to differences between foundations, particularly those individualizing and binding, people may not always recognize each other's values as moral~\cite{haidtWhenMoralityOpposes2007}.
Yet, endorsement of binding foundations doesn't exclude praising individualizing foundations and vice versa. In other words, the moral compass of an individual doesn't point straight in one direction.

\begin{figure}
\includegraphics[width=0.95\linewidth]{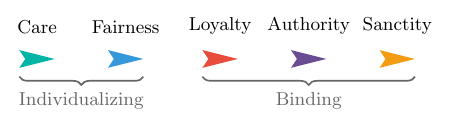}
\caption{Foundations: from individualism to collectivism.}
\end{figure}

\subsection{Measuring Moral Foundations}

Moral questionnaires, such as the Moral Foundations Questionnaire, are
available in multiple languages~\cite{grahamMappingMoralDomain2011,atariMoralityWEIRDHow2023}.
Self-report is well suited to profiling individuals and comparing
groups, but it's not applicable to the kind of naturally occurring text
--- social media posts and other everyday expression --- that text
analytics relies on, where no questionnaire response is available to
draw on.

Dictionary-based methods~\cite{araque_moralstrength_2020,hoppExtendedMoralFoundations2021}
assess moral content via lexicon matching to foundation-specific
keywords and phrases, but are imprecise in capturing context and
nuance, as discussed in~\cite{kennedy_moral_2023}.

Supervised classification is possible directly on natural text, but
requires annotated corpora~\cite{hooverMoralFoundationsTwitter2020,tragerMoralFoundationsReddit2022}
that exist almost exclusively in English; extending to new languages
demands substantial re-annotation effort, with awareness of both
MFT and language-specific subtleties.

\subsection{Cross-Lingual Transfer and Translation}

Two zero-shot strategies could, in principle, avoid corpus translation
altogether. Cross-lingual encoders such as mBERT~\cite{devlin2019bert}
and XLM-RoBERTa~\cite{conneau2020unsupervised} enable zero-shot
transfer to new languages without additional annotation; however,
without fine-tuning, they introduce systematic biases that distort
cross-lingual text analysis and affective content~\cite{nicholas2023lost,plazadelarco2024angry}.
Alternatively, an LLM could classify moral content directly in the
target language, skipping translation entirely; however, LLMs
themselves exhibit inconsistent moral reasoning across
languages~\cite{aksoyWhoseMoralityThey2025}, undermining this route too.

Prior cross-lingual moral analysis has instead relied on multilingual
dictionaries to map moral seed words across languages~\cite{hoppExtendedMoralFoundations2021,matsuo_development_2019}, without exploiting
machine translation at scale.

For our own validation, LaBSE~\cite{feng2020language} provides strong
cross-lingual sentence embeddings well-suited for measuring semantic
equivalence between source and translated text --- the tool we adopt to assess translation fidelity. 
We instead translate the full corpus with an LLM and fine-tune a classifier on
the translated data: this combines the scalability of LLM translation
with the robustness of fine-tuning, avoiding both the bias risks of
the zero-shot approaches and the limited scale of dictionary-based
methods.

\section{Methods}

\begin{figure}[t]
\centering
\includegraphics[width=0.99\linewidth]{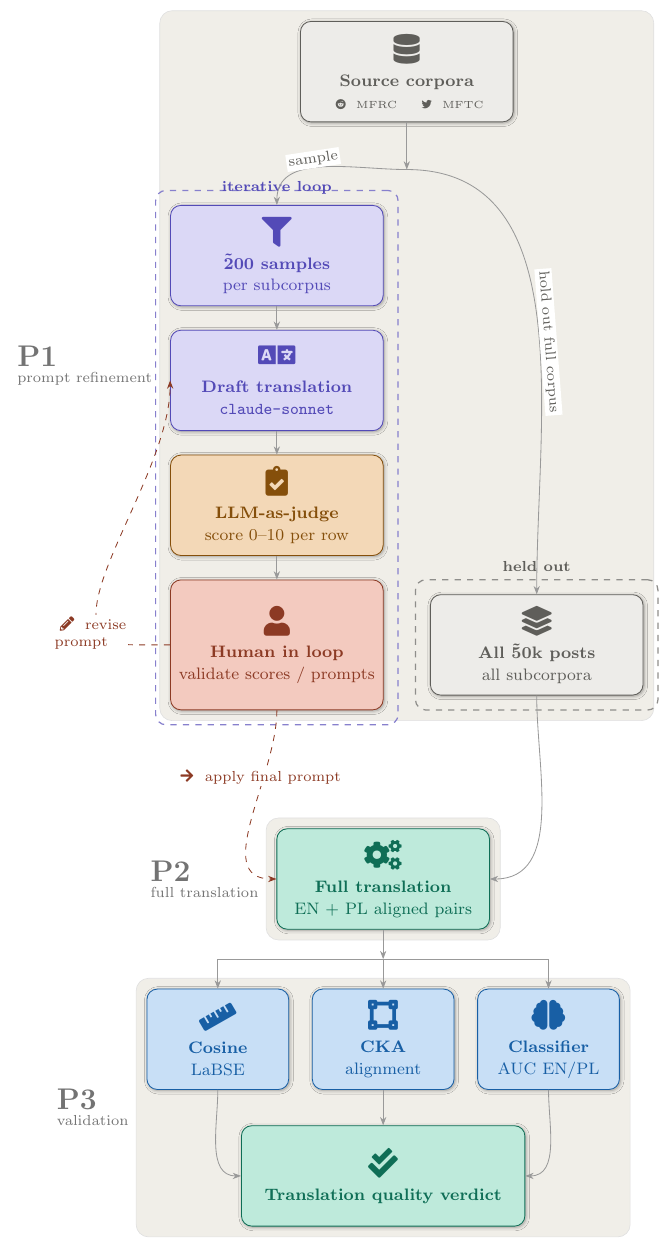}
\Description{Diagram of the three-phase validation pipeline: prompt refinement, full corpus translation, and fidelity evaluation.}
\caption{Validation pipeline. Phase~1 iteratively refines the translation
prompt on subcorpus samples via an LLM judge coupled with human-in-loop, while the full \textasciitilde50k-post
corpus is held out. Phase~2 translates the full corpus, producing aligned EN/PL pairs.
Phase~3 evaluates fidelity via cosine similarity, CKA, and classifier performance
AUROC.}
\label{fig:pipeline}
\end{figure}

\subsection{Corpora}

Two corpora were selected to cover a variety of moral discourse
styles and difficulty levels (see~\Cref{tab:corpora}).

\textbf{MFRC} (Moral Foundations Reddit Corpus)~\cite{tragerMoralFoundationsReddit2022}
provides 17{,}886 Reddit posts labeled across five MFT foundations
(authority, care, fairness, loyalty, sanctity) spanning three
subcorpora: everyday morality (\textit{r/AmItheAsshole}), US
politics, and French politics. The corpus covers moral reasoning
expressed through colloquial language, abbreviations (NTA, YTA,
AITA), and informal style.

\textbf{MFTC} (Moral Foundations Twitter Corpus)~\cite{hooverMoralFoundationsTwitter2020}
provides 33{,}858 Twitter posts across seven subcorpora --- spanning
political movements (\#BLM, \#MeToo, All Lives Matter), civil
unrest (Baltimore uprising), electoral discourse, and a natural
disaster (Hurricane Sandy). The Davidson subcorpus, drawn from a
hate speech dataset~\cite{davidson2017automated}, represents a
hard test case with vulgarity and AAVE-heavy language.

Both corpora were hand-annotated by multiple trained annotators
per post~\cite{hooverMoralFoundationsTwitter2020,tragerMoralFoundationsReddit2022}
using the same five foundations; a substantial fraction of posts in
both corpora carry no moral label at all, serving as the negative
class for per-foundation classification.

These data cover a diverse range of moral discourse:
everyday interpersonal judgments, heated political debate, social
movements, and collective responses to crisis --- making them a
demanding and representative benchmark for cross-lingual translation.

\begin{table*}[h!]
  \caption{Corpora, subcorpora, and per-foundation prevalence (text flagged at least once by any annotator, as \% of total texts). Abbreviations:
           Au=authority, Ca=care, Fa=fairness, Lo=loyalty, Sa=sanctity.
           Totals include non-moral instances.}
  \label{tab:corpora}
  \setlength{\tabcolsep}{5pt}
  \begin{tabular}{llllrrrrrrr}
    \toprule
    Corpus & Subcorpus & Platform & Domain
      & Au\% & Ca\% & Fa\% & Lo\% & Sa\% & $N$ \\
    \midrule
    \multirow{4}{*}{MFRC}
      & Everyday morality & Reddit & General moral discourse
        & 10.4 & 37.4 & 25.5 & 11.7 & 13.5 &  5{,}366 \\
      & US politics       & Reddit & US political discourse
        & 19.7 & 29.6 & 38.3 &  7.7 &  8.4 &  5{,}351 \\
      & French politics   & Reddit & French political discourse
        & 25.4 & 16.0 & 26.0 & 13.1 &  8.0 &  7{,}169 \\
    \cmidrule{2-10}
      & \multicolumn{4}{r}{\textit{MFRC total}} & & & & & & 17{,}886 \\
    \midrule
    \multirow{8}{*}{MFTC}
      & ALM       & Twitter & All Lives Matter
        & 20.9 & 61.6 & 49.8 & 26.3 & 16.9 &  4{,}326 \\
      & BLM       & Twitter & Black Lives Matter
        & 21.3 & 56.6 & 49.5 & 27.2 & 20.8 &  5{,}117 \\
      & Baltimore & Twitter & Baltimore uprising
        & 31.8 & 27.1 & 31.4 & 42.3 &  9.0 &  5{,}190 \\
      & Davidson  & Twitter & Hate speech
        & 33.0 & 11.5 & 11.2 & 11.6 & 12.3 &  4{,}873 \\
      & Election  & Twitter & US election
        & 17.5 & 38.0 & 34.4 & 22.8 & 30.4 &  5{,}050 \\
      & MeToo     & Twitter & MeToo movement
        & 65.7 & 33.3 & 43.9 & 41.0 & 52.8 &  4{,}711 \\
      & Sandy     & Twitter & Hurricane Sandy
        & 45.6 & 60.9 & 30.3 & 42.7 & 15.0 &  4{,}591 \\
    \cmidrule{2-10}
      & \multicolumn{4}{r}{\textit{MFTC total}} & & & & & & 33{,}858 \\
    \bottomrule
  \end{tabular}
\end{table*}

\subsection{Translation Pipeline}

Translation was performed using
\texttt{Claude-Sonnet-4-6} via the Anthropic
API, with 20 concurrent asynchronous requests to maximise throughput.
Platform-specific prompting was applied (~\Cref{sec:prompts}): the Reddit prompt instructs the model to preserve informal tone,
Reddit abbreviations (NTA, YTA), and formatting; the Twitter prompt
additionally preserves hashtags and @mentions unchanged.
The full translation cost approximately 200~USD for 50k posts combined,
demonstrating the accessibility of this approach for research groups
without large annotation budgets.

\subsection{Translation Prompts}\label{sec:prompts}

Two platform-specific system prompts were carefully engineered to
handle the distinct linguistic styles of each corpus
(Prompts~P1--P2). Both share a common design philosophy: preserve
moral-semantic content while naturalizing tone into idiomatic Polish.
Key design decisions include explicit slang mappings
(e.g.\ \textit{wtf}$\to$\textit{kurwa}), strict hashtag and mention
preservation, and grammar rules for name declension.  To account for the differences between Reddit and Twitter, the prompts differ in their handling of Reddit-specific conventions (NTA/YTA abbreviations, Markdown formatting, nested quotes) versus Twitter-specific ones (ALL CAPS emphasis, retweet abbreviations, activist hashtags); a single merged prompt risked cross-applying rules across platforms (e.g., treating Reddit quote markers as hashtags), so we kept them separate. Each prompt includes a one-shot example to anchor the target style. Both prompts were iteratively refined
using a stratified sample of \textasciitilde200 posts per subcorpus,
evaluated by an LLM judge on tone preservation, slang handling,
formatting fidelity, and proper noun treatment --- as detailed in
\Cref{sec:validation}.

\begin{center}
\includegraphics[width=0.99\linewidth]{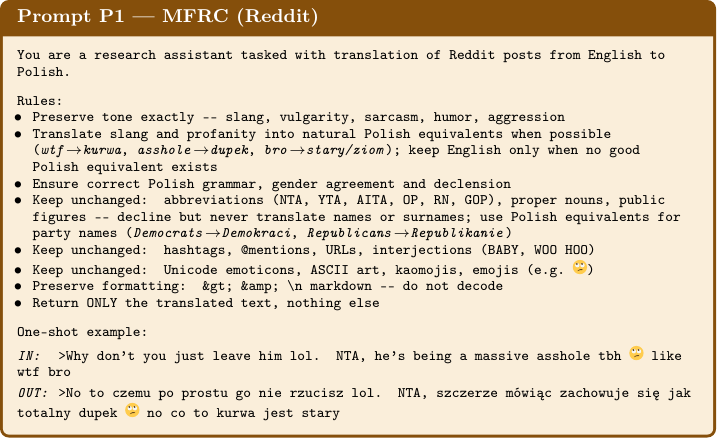}
\Description{Screenshot of the system prompt used to translate Reddit posts, with rules for tone, slang, and formatting.}
\end{center}

\begin{center}
\includegraphics[width=0.99\linewidth]{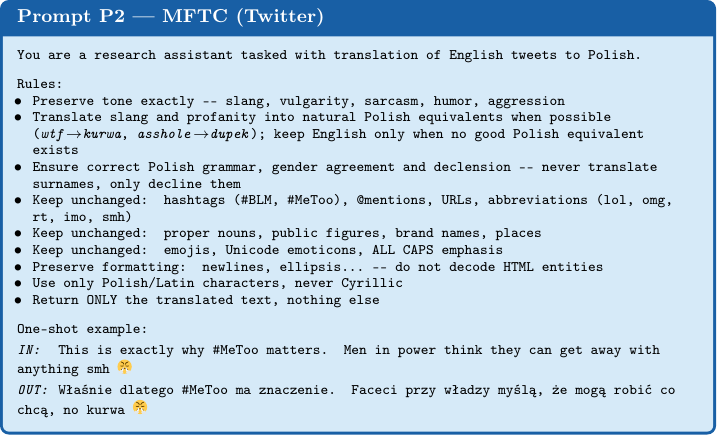}
\Description{Screenshot of the system prompt used to translate tweets, with rules for hashtags, mentions, and tone.}
\end{center}

\subsection{Validation Methods}\label{sec:validation}

The validation pipeline proceeds in three phases
(\Cref{fig:pipeline}). Phase~1 uses \textasciitilde200 samples
per subcorpus to iteratively refine translation prompts via an LLM
judge. Phase~2 applies the final prompt to all subcorpora. Finally, Phase~3
evaluates translation fidelity through four complementary methods.

\textbf{Embedding Similarity (LaBSE).} Cross-lingual cosine
similarity between English and Polish sentence pairs was computed
using LaBSE~\cite{feng2020language}. The expected range for
well-translated pairs is 0.80--0.95, consistent with prior findings
in the literature. The work~\cite{kumari_quality_2024}
find that nearly all sentence pairs with LaBSE similarity $\geq 0.80$
also score highly on an independent, corpus-derived quality metric, on
a large-scale English--Hindi corpus. Similarly,~\cite{steingrimsson_sentence_2023} accepts sentence
pairs with LaBSE similarity score above 0.9 directly into training pipelines
without further quality filtering.

\textbf{Centered Kernel Alignment (CKA).} CKA~\cite{kornblith2019similarity} measures global alignment between embedding spaces, providing a
stronger signal than mean pairwise similarity, as it captures structural
preservation beyond individual sentence pairs. CKA compares similarity structure across the full set of embeddings
using a linear kernel~\cite{kornblith2019similarity} and is invariant
to orthogonal rotation and isotropic scaling of either embedding
space. Prior work on cross-lingual parallel
sentence representations found CKA similarity to highly correlate with
sentence-retrieval alignment accuracy~\cite{conneau_emerging_2020}, supporting its use here as a
reliable signal of cross-lingual semantic preservation.

\textbf{LLM-as-Judge.} Translation quality is increasingly judged by
LLMs directly, without a reference translation; it was demonstrated
that GPT-class models match human quality ratings from WMT, the main
machine translation benchmark, as closely as established automatic
metrics do~\cite{kocmi_large_2023}. In line with this, a stratified
sample of \textasciitilde200 posts per subcorpus was evaluated by
Claude Sonnet on four dimensions: tone preservation, slang handling,
formatting fidelity, and proper noun treatment. Scores were elicited
on a 0--10 scale.

\textbf{Classifier Parity / Gap Validation.} A linear classification
head was trained on frozen LaBSE embeddings using 10-fold stratified
cross-validation; this is a standard technique for testing what a
representation already contains, since a simple classifier on frozen
features can't rely on extra learned capacity to make up for weak
embeddings~\cite{alain_understanding_2017}. Out-of-fold predicted
probabilities were pooled across folds and compared between English
and Polish using DeLong's test \cite{delong_comparing_1988,sun_fast_2014}
on the correlated ROC-AUCs, giving a two-sided $p$-value and an exact
95\% confidence interval for the AUC gap; we report this at both the
corpus and subcorpus levels. Frozen-embedding AUC
is intentionally conservative: full fine-tuning lifts both languages'
performance jointly, so the reported gaps isolate translation fidelity
rather than absolute classifier quality. As a supplementary check,
mDeBERTa-v3-base~\cite{he_debertav3_2023} was fully fine-tuned
end-to-end on both English and translated Polish corpora to confirm
parity holds under full gradient updates. This model was chosen
because it was recently shown to outperform other BERT-family models
on MFT classification~\cite{skorskiMoralGapLarge2025}.


\section{Results}

\subsection{Model-as-Judge Quality}

Translation quality was assessed via row-by-row LLM-as-judge
evaluation ($N{\approx}200$ per subcorpus) on a 0--10 scale
(~\Cref{tab:judge-audit}). Across all subcorpora the mean score
is \textbf{8.9}, with 94.6\% of posts free of detectable issues.
Scores are lowest on five subcorpora (8.5): three where AAVE and
dialect-heavy language force paraphrase (ALM, Davidson, Baltimore),
and two where difficulty stems from register and topic complexity
rather than dialect (Everyday Morality, French Politics). Scores are
highest on BLM and Sandy (9.5). Two model-level failure modes persist regardless of prompt engineering: sporadic hashtag content
translation and Cyrillic character leakage during self-correction. 
Overall, the scores are consistent with the state-of-the-art judge reliability reported for GPT-class evaluators against WMT human ratings~\cite{kocmi_large_2023}.

\begin{table}[h!]
\centering
\caption{Row-by-row LLM-judge translation audit (EN$\to$PL, $N{=}200$ per sub-corpus).
\textit{Clean}: no issues. \textit{Minor}: tone softening, inconsistent slang, formatting artefacts.
\textit{Errors}: grammar failures, meaning inversions, untranslated segments, spurious refusals.
\textit{Score}: 0--10 judgment (human validated).}
\label{tab:judge-audit}
\small
\setlength{\tabcolsep}{4pt}
\begin{tabular}{llrrrc}
\toprule
\textbf{Corpus} & \textbf{Sub-corpus} & \textbf{Clean\,\%} & \textbf{Minor\,\%} & \textbf{Err.\,\%} & \textbf{Score} \\
\midrule
\multirow{3}{*}{MFRC}
  & Everyday Morality & 93.0 & 5.0 & 2.0 & 8.5 \\
  & US Politics       & 95.0 & 3.0 & 2.0 & 9.5 \\
  & French Politics   & 93.0 & 5.0 & 2.0 & 8.5 \\
\midrule
\multirow{7}{*}{MFTC}
  & ALM               & 91.0 & 7.0 & 2.0 & 8.5 \\
  & BLM               & 95.5 & 3.5 & 1.0 & 9.5 \\
  & Baltimore         & 95.0 & 3.5 & 1.5 & 8.5 \\
  & Davidson          & 94.0 & 4.5 & 1.5 & 8.5 \\
  & Election          & 96.5 & 2.5 & 1.0 & 9.0 \\
  & MeToo             & 96.5 & 2.5 & 1.0 & 9.0 \\
  & Sandy             & 96.5 & 2.5 & 1.0 & 9.5 \\
\midrule
\multicolumn{2}{l}{\textbf{Average}} & \textbf{94.6} & \textbf{3.9} & \textbf{1.5} & \textbf{8.9} \\
\bottomrule
\end{tabular}
\end{table}

\subsection{Embedding Similarity and CKA}

LaBSE cross-lingual cosine similarity and linear CKA are reported in
\Cref{tab:cosine,tab:cka}. Mean cosine similarity is
\textbf{0.889} overall (MFRC: 0.876, MFTC: 0.894), well above the
random baseline of ${\approx}0.30$ and exceeding the 0.80 threshold
considered strong semantic equivalence. CKA confirms good global embedding alignment (overall \textbf{0.860}),
with French Politics and Baltimore scoring highest (0.895--0.896)
and Davidson lowest (0.806), where AAVE paraphrase shifts
distributional geometry beyond what pairwise distances capture.
A modest gap to 1.0 is expected, attributable to Polish
morphological inflection and culture-specific expressions.
Overall, this indicates high fidelity of the translation, as per the recent literature~\cite{kumari_quality_2024,steingrimsson_sentence_2023}.

\begin{table}[h!]
\caption{LaBSE cross-lingual cosine similarity between English source and Polish
translation per sub-corpus. P05/P95 denote the 5th and 95th percentiles.
A threshold of ${\geq}0.80$ is widely considered strong semantic equivalence.}
\label{tab:cosine}
\centering
\small
\setlength{\tabcolsep}{4pt}
\begin{tabular}{llrrrrr}
\toprule
\textbf{Corpus} & \textbf{Sub-corpus} & \textbf{N} & \textbf{Mean} & \textbf{Std} & \textbf{P05} & \textbf{P95} \\
\midrule
\multirow{3}{*}{MFRC}
  & Everyday Morality & 5{,}366 & 0.869 & 0.052 & 0.782 & 0.929 \\
  & US Politics       & 5{,}351 & 0.863 & 0.056 & 0.773 & 0.926 \\
  & French Politics   & 7{,}169 & 0.896 & 0.039 & 0.831 & 0.946 \\
\midrule
\multirow{7}{*}{MFTC}
  & ALM               & 4{,}326 & 0.911 & 0.042 & 0.834 & 0.970 \\
  & BLM               & 5{,}117 & 0.921 & 0.047 & 0.840 & 0.985 \\
  & Baltimore         & 5{,}190 & 0.911 & 0.063 & 0.809 & 1.000 \\
  & Davidson          & 4{,}873 & 0.867 & 0.083 & 0.712 & 0.962 \\
  & Election          & 5{,}050 & 0.901 & 0.056 & 0.808 & 0.967 \\
  & MeToo             & 4{,}711 & 0.903 & 0.052 & 0.820 & 0.979 \\
  & Sandy             & 4{,}591 & 0.843 & 0.100 & 0.713 & 0.937 \\
\midrule
\multicolumn{2}{l}{\textbf{Overall}} & \textbf{51{,}744} & \textbf{0.889} & \textbf{0.063} & \textbf{0.789} & \textbf{0.960} \\
\bottomrule
\end{tabular}
\end{table}

\begin{table}[h!]
\caption{Linear CKA between LaBSE embeddings of English source and Polish translation
per sub-corpus. CKA measures global alignment of embedding spaces; a value of 1.0
indicates identical geometry up to orthogonal transformation.}
\label{tab:cka}
\centering
\small
\setlength{\tabcolsep}{5pt}
\begin{tabular}{llrr}
\toprule
\textbf{Corpus} & \textbf{Sub-corpus} & \textbf{N} & \textbf{CKA} \\
\midrule
\multirow{3}{*}{MFRC}
  & Everyday Morality & 5{,}366 & 0.833 \\
  & US Politics       & 5{,}351 & 0.826 \\
  & French Politics   & 7{,}169 & 0.895 \\
\midrule
\multirow{7}{*}{MFTC}
  & ALM               & 4{,}326 & 0.867 \\
  & BLM               & 5{,}117 & 0.883 \\
  & Baltimore         & 5{,}190 & 0.896 \\
  & Davidson          & 4{,}873 & 0.806 \\
  & Election          & 5{,}050 & 0.877 \\
  & MeToo             & 4{,}711 & 0.836 \\
  & Sandy             & 4{,}591 & 0.845 \\
\midrule
\multicolumn{2}{l}{\textbf{Overall}} & \textbf{51{,}744} & \textbf{0.860} \\
\bottomrule
\end{tabular}
\end{table}

\subsection{Qualitative Translation Analysis}\label{sec:qual-error}

Two common failure modes surfaced on manual review of the translated corpus:
\textcolor{clrAmberDark}{\textbf{morphological}} errors (case/gender
agreement --- declension mistakes typical of Polish's seven-case
system) and \textcolor{clrPurpleDark}{\textbf{tonal}} distortion
(grammatically correct slang substitution that shifts register
relative to the English source). A correctly naturalized example is included as a reference point. 

\begin{figure}
\includegraphics[width=0.99\linewidth]{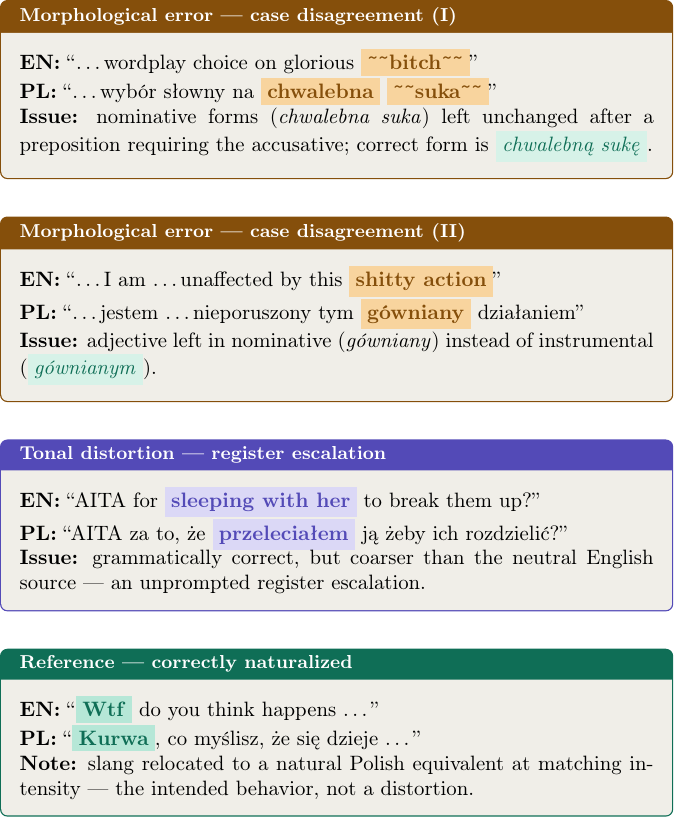}
\Description{Example English-Polish translation pairs showing correct handling of slang and morphology, plus common error cases.}
\end{figure}

\subsection{Classifiers Parity}





\Cref{tab:parity} reports AUROC scores for classification per foundation, using a linear classifier trained on top of LaBSE embeddings (10-fold cross-validation). 

At the corpus level, DeLong's test~\cite{delong_comparing_1988,sun_fast_2014} on pooled out-of-fold predictions (\Cref{tab:parity}, ``All (pooled)'' rows) gives exact 95\% CIs for the EN--PL score gap: these fall below the 0.02 threshold for care, fairness, loyalty, and sanctity in both corpora, and only marginally exceed it for authority (MFRC $0.011$--$0.021$; MFTC $0.016$--$0.022$).

Across all 50 subcorpus-foundation pairs, the classification performance gaps between English and Polish remain small, with median of \textbf{0.018} and inter-quartile range \textbf{0.009--0.023}.

On MFRC, most gaps between English and Polish versions are below 0.015 --- fairness and sanctity show no significant degradation.
Authority shows the highest gaps (0.021--0.032) of the five foundations, consistent with the cross-cultural divergence discussed below.

On MFTC, gaps are larger on AAVE-heavy subcorpora (ALM, Election, Sandy: up to 0.047) yet no foundation is invalidated for downstream use. Davidson is a special case: near-chance baseline AUC reflects weak moral signal in hate speech, not translation failure --- care and fairness even show negative gaps (PL > EN).

Frozen-embedding AUROC is intentionally conservative: full fine-tuning lifts both EN and PL jointly (as demonstrated in \Cref{sec:finetuning}), so gaps here measure translation fidelity, not absolute classifier quality~\cite{alain_understanding_2017}.

\begin{table*}[h!]
  \caption{Classifier parity: ROC-AUC per moral foundation and subcorpus (EN vs.\ PL),
           linear head on frozen LaBSE embeddings, DeLong's test on pooled out-of-fold predictions.
           \textit{All (pooled)} rows combine all subcorpora within the corpus.
           $p_{\text{diff}}$: two-sided test for EN/PL AUC gap, and 95\% confidence interval both estimated from DeLong's test.}
  \label{tab:parity}
  \setlength{\tabcolsep}{4pt}
  \footnotesize
  \begin{tabular}{lllcccccc}
    \toprule
    Corpus & Subcorpus & Foundation & EN AUC & PL AUC & Gap & SE & $p_{\text{diff}}$ & 95\% CI \\
    \midrule
    \multirow{20}{*}{MFRC}
          & \multirow{5}{*}{\textbf{All (pooled)}} & \textbf{Authority} & \textbf{0.801} & \textbf{0.785} & $+$0.016 & 0.003 & $<$0.001 & 0.011--0.021 \\
        && \textbf{Care} & \textbf{0.850} & \textbf{0.843} & $+$0.007 & 0.002 & $<$0.001 & 0.003--0.011 \\
        && \textbf{Fairness} & \textbf{0.776} & \textbf{0.771} & $+$0.006 & 0.002 & 0.011 & 0.001--0.011 \\
        && \textbf{Loyalty} & \textbf{0.785} & \textbf{0.775} & $+$0.010 & 0.004 & 0.009 & 0.003--0.017 \\
        && \textbf{Sanctity} & \textbf{0.766} & \textbf{0.765} & $+$0.001 & 0.004 & 0.774 & -0.007--0.009 \\
    \cmidrule{2-9}
          & \multirow{5}{*}{Everyday} & Authority & 0.813 & 0.787 & $+$0.026 & 0.008 & 0.001 & 0.011--0.041 \\
        && Care & 0.853 & 0.842 & $+$0.011 & 0.004 & 0.004 & 0.004--0.018 \\
        && Fairness & 0.790 & 0.784 & $+$0.007 & 0.005 & 0.208 & -0.003--0.017 \\
        && Loyalty & 0.802 & 0.793 & $+$0.009 & 0.008 & 0.237 & -0.006--0.024 \\
        && Sanctity & 0.728 & 0.725 & $+$0.004 & 0.009 & 0.675 & -0.013--0.021 \\
    \cmidrule{2-9}
          & \multirow{5}{*}{US politics} & Authority & 0.791 & 0.759 & $+$0.032 & 0.006 & $<$0.001 & 0.019--0.045 \\
        && Care & 0.767 & 0.753 & $+$0.014 & 0.005 & 0.007 & 0.004--0.024 \\
        && Fairness & 0.712 & 0.707 & $+$0.005 & 0.006 & 0.338 & -0.006--0.016 \\
        && Loyalty & 0.757 & 0.735 & $+$0.022 & 0.010 & 0.034 & 0.002--0.042 \\
        && Sanctity & 0.698 & 0.707 & $-$0.010 & 0.012 & 0.408 & -0.033--0.013 \\
    \cmidrule{2-9}
          & \multirow{5}{*}{French politics} & Authority & 0.727 & 0.706 & $+$0.021 & 0.005 & $<$0.001 & 0.011--0.031 \\
        && Care & 0.814 & 0.794 & $+$0.020 & 0.005 & $<$0.001 & 0.010--0.030 \\
        && Fairness & 0.726 & 0.718 & $+$0.008 & 0.005 & 0.079 & -0.001--0.017 \\
        && Loyalty & 0.720 & 0.698 & $+$0.022 & 0.007 & 0.003 & 0.007--0.037 \\
        && Sanctity & 0.741 & 0.738 & $+$0.004 & 0.008 & 0.659 & -0.012--0.020 \\
    \midrule
    \multirow{40}{*}{MFTC}
          & \multirow{5}{*}{\textbf{All (pooled)}} & \textbf{Authority} & \textbf{0.837} & \textbf{0.818} & $+$0.019 & 0.001 & $<$0.001 & 0.016--0.022 \\
        && \textbf{Care} & \textbf{0.845} & \textbf{0.835} & $+$0.010 & 0.001 & $<$0.001 & 0.008--0.012 \\
        && \textbf{Fairness} & \textbf{0.843} & \textbf{0.830} & $+$0.013 & 0.001 & $<$0.001 & 0.011--0.015 \\
        && \textbf{Loyalty} & \textbf{0.806} & \textbf{0.792} & $+$0.014 & 0.001 & $<$0.001 & 0.011--0.017 \\
        && \textbf{Sanctity} & \textbf{0.813} & \textbf{0.802} & $+$0.011 & 0.002 & $<$0.001 & 0.008--0.014 \\
    \cmidrule{2-9}
          & \multirow{5}{*}{ALM} & Authority & 0.875 & 0.828 & $+$0.047 & 0.006 & $<$0.001 & 0.036--0.058 \\
        && Care & 0.795 & 0.772 & $+$0.022 & 0.005 & $<$0.001 & 0.012--0.032 \\
        && Fairness & 0.788 & 0.757 & $+$0.031 & 0.005 & $<$0.001 & 0.020--0.042 \\
        && Loyalty & 0.738 & 0.735 & $+$0.003 & 0.007 & 0.605 & -0.010--0.016 \\
        && Sanctity & 0.714 & 0.707 & $+$0.007 & 0.009 & 0.453 & -0.010--0.024 \\
    \cmidrule{2-9}
          & \multirow{5}{*}{BLM} & Authority & 0.889 & 0.862 & $+$0.027 & 0.005 & $<$0.001 & 0.017--0.037 \\
        && Care & 0.834 & 0.820 & $+$0.014 & 0.004 & $<$0.001 & 0.006--0.022 \\
        && Fairness & 0.813 & 0.791 & $+$0.022 & 0.004 & $<$0.001 & 0.014--0.030 \\
        && Loyalty & 0.809 & 0.788 & $+$0.021 & 0.005 & $<$0.001 & 0.011--0.031 \\
        && Sanctity & 0.825 & 0.816 & $+$0.009 & 0.005 & 0.070 & -0.001--0.019 \\
    \cmidrule{2-9}
          & \multirow{5}{*}{Baltimore} & Authority & 0.857 & 0.846 & $+$0.012 & 0.003 & $<$0.001 & 0.005--0.019 \\
        && Care & 0.815 & 0.801 & $+$0.014 & 0.004 & 0.002 & 0.005--0.023 \\
        && Fairness & 0.867 & 0.857 & $+$0.010 & 0.003 & 0.002 & 0.004--0.016 \\
        && Loyalty & 0.880 & 0.869 & $+$0.011 & 0.003 & $<$0.001 & 0.005--0.017 \\
        && Sanctity & 0.758 & 0.740 & $+$0.018 & 0.009 & 0.039 & 0.001--0.035 \\
    \cmidrule{2-9}
          & \multirow{5}{*}{Davidson} & Authority & 0.596 & 0.596 & $+$0.000 & 0.008 & 0.997 & -0.015--0.015 \\
        && Care & 0.506 & 0.510 & $-$0.004 & 0.014 & 0.768 & -0.031--0.023 \\
        && Fairness & 0.549 & 0.552 & $-$0.002 & 0.013 & 0.847 & -0.027--0.023 \\
        && Loyalty & 0.596 & 0.575 & $+$0.021 & 0.012 & 0.070 & -0.002--0.044 \\
        && Sanctity & 0.591 & 0.570 & $+$0.021 & 0.012 & 0.073 & -0.002--0.044 \\
    \cmidrule{2-9}
          & \multirow{5}{*}{Election} & Authority & 0.838 & 0.801 & $+$0.037 & 0.006 & $<$0.001 & 0.025--0.049 \\
        && Care & 0.800 & 0.782 & $+$0.018 & 0.005 & $<$0.001 & 0.009--0.027 \\
        && Fairness & 0.864 & 0.834 & $+$0.029 & 0.004 & $<$0.001 & 0.021--0.037 \\
        && Loyalty & 0.753 & 0.735 & $+$0.018 & 0.006 & 0.003 & 0.006--0.030 \\
        && Sanctity & 0.828 & 0.817 & $+$0.011 & 0.005 & 0.018 & 0.002--0.020 \\
    \cmidrule{2-9}
          & \multirow{5}{*}{MeToo} & Authority & 0.764 & 0.761 & $+$0.003 & 0.005 & 0.587 & -0.008--0.014 \\
        && Care & 0.824 & 0.809 & $+$0.015 & 0.005 & 0.004 & 0.005--0.025 \\
        && Fairness & 0.806 & 0.787 & $+$0.019 & 0.005 & $<$0.001 & 0.009--0.029 \\
        && Loyalty & 0.761 & 0.740 & $+$0.020 & 0.006 & $<$0.001 & 0.009--0.031 \\
        && Sanctity & 0.760 & 0.744 & $+$0.016 & 0.005 & 0.004 & 0.005--0.027 \\
    \cmidrule{2-9}
          & \multirow{5}{*}{Sandy} & Authority & 0.883 & 0.858 & $+$0.025 & 0.004 & $<$0.001 & 0.017--0.033 \\
        && Care & 0.846 & 0.820 & $+$0.026 & 0.005 & $<$0.001 & 0.017--0.035 \\
        && Fairness & 0.865 & 0.827 & $+$0.038 & 0.005 & $<$0.001 & 0.028--0.048 \\
        && Loyalty & 0.802 & 0.778 & $+$0.024 & 0.005 & $<$0.001 & 0.014--0.034 \\
        && Sanctity & 0.804 & 0.762 & $+$0.042 & 0.009 & $<$0.001 & 0.025--0.059 \\
    \bottomrule
  \end{tabular}
\end{table*}

\begin{figure}[h!]
  \centering
  \begin{subfigure}[b]{0.99\columnwidth}
    \includegraphics[width=\linewidth]{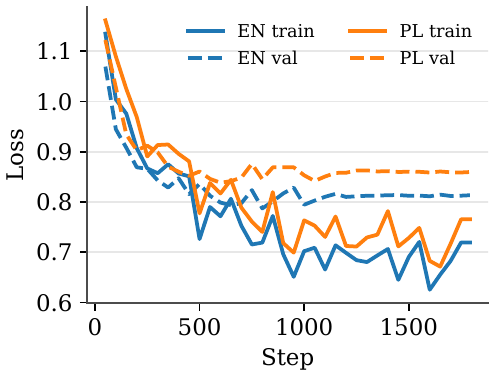}
    \Description{Training loss curves for English and Polish, nearly overlapping throughout training.}
    \caption{Loss}
    \label{fig:loss-curves_a}
  \end{subfigure}

  \begin{subfigure}[b]{0.99\columnwidth}
    \includegraphics[width=\linewidth]{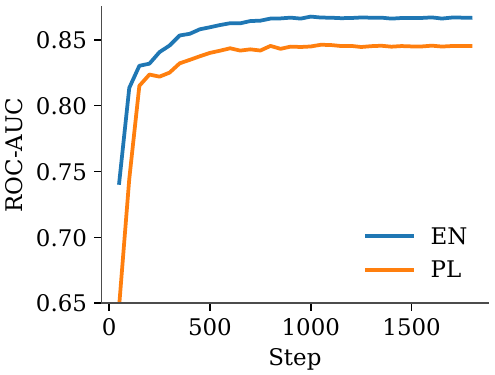}
    \caption{ROC-AUC}
    \Description{ROC-AUC curves for English and Polish, converging to nearly the same value.}
    \label{fig:auc-curves_a}
  \end{subfigure}
  \caption{Full fine-tuning on MFRC Care (mDeBERTa-v3, lr=5e-5).
           EN and PL on near-identical learning curves.}
  \label{fig:finetuning_a}
\end{figure}

\begin{figure}[h!]
  \centering
  \begin{subfigure}[b]{0.99\columnwidth}
    \includegraphics[width=\linewidth]{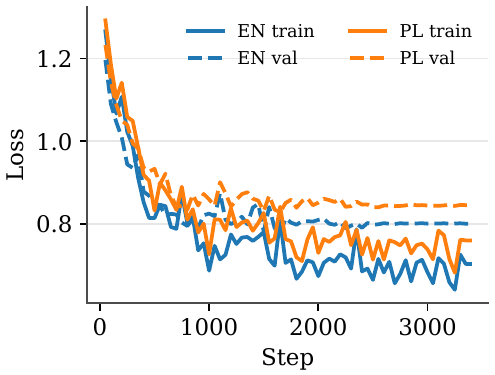}
    \Description{ROC-AUC curves for English and Polish on the Davidson subcorpus, converging to a small final gap.}
    \caption{Loss}
    \label{fig:loss-curves_b}
  \end{subfigure}

  \begin{subfigure}[b]{0.99\columnwidth}
    \includegraphics[width=\linewidth]{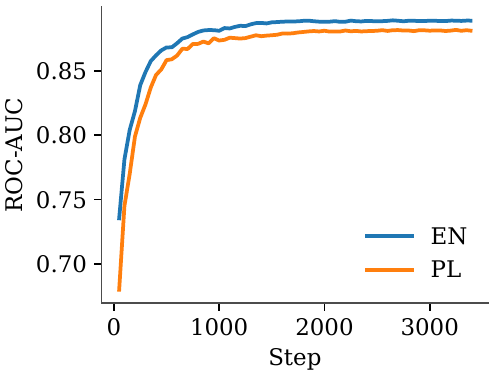}
    \Description{Five colored arrows for the moral foundations, grouped into individualizing (care, fairness) and binding (loyalty, authority, sanctity).}
    \caption{ROC-AUC}
    \label{fig:auc-curves_b}
  \end{subfigure}
  \caption{Full fine-tuning on MFTC Fairness (mDeBERTa-v3, lr=5e-5).
           EN and PL on near-identical learning curves.}
  \label{fig:finetuning_b}
\end{figure}

\subsection{Full Fine-Tuning Validation}\label{sec:finetuning}

mDeBERTa-v3-base was fully fine-tuned on both English and translated
Polish corpora, to confirm near-parity also under full gradient updates.
As illustrated in \Cref{fig:finetuning_a} and \Cref{fig:finetuning_b} for MFRC Care and MFTC Fairness,
EN and PL trainings follow near-identical trajectories,
with a final AUC gap smaller than \textbf{0.02}. On the harder MFTC Davidson
subcorpus (fairness; near-chance frozen baseline, AAVE-heavy language),
trajectories remain parallel, converging from $\approx$0.57 to
$\approx$0.68 with a final tiny gap of \textbf{0.006}. More results are available in the preprint version of this paper.

\section{Discussion}

\textbf{Overall.}
The convergent evidence from four independent validation methods
supports a perhaps surprising conclusion: LLM-based EN$\to$PL
translation preserves moral-semantic content at a level sufficient
for downstream classification. Moral language --- with its irony,
idiom, and cultural sensitivity --- proves more robust to
translation than research on LLM limitations with regard to affective signals might
suggest~\cite{nicholas2023lost,plazadelarco2024angry}. 

\textbf{Authority gap.} Among the five foundations, the AUC gap on authority remains largest, with its 95\% CI slightly exceeding 0.02 (\Cref{tab:parity}). We attribute this not to translation error but to \textbf{genuine cross-cultural divergence in how authority is
expressed in Polish discourse} --- an observation consistent with the
cross-cultural MFT literature~\cite{graham2013moral,skorskiMoralGapLarge2025},
where authority norms show the highest cross-national variance and the
highest sensitivity to domain variation. Future work could verify this by
re-annotating a Polish sample with native speakers trained in MFT.

\textbf{Harder corpora.} Difficulty tracks dialect and register, not platform: the lowest scores cluster in AAVE-heavy subcorpora regardless of source --- Davidson (CKA 0.806, judge 8.5) and ALM (CKA 0.867, judge 8.5) in MFTC, alongside Everyday Morality and French Politics (judge 8.5) in MFRC --- where translation necessarily involves paraphrase rather than literal mapping. Corpus-level averages are in fact comparable (CKA: MFRC 0.851, MFTC 0.859; judge: MFRC 8.8, MFTC 8.9). Even so, classifier parity confirms the translated data remains usable for training across all subcorpora.

\textbf{Practical accessibility.}
The pipeline is accessible: $\sim$50k posts translate for
approximately 200~USD, making corpus extension to additional
languages feasible without large annotation
budgets.

\textbf{Other Slavic languages.}
Polish is among the most morphologically complex languages in the
Slavic family, with seven grammatical cases and rich inflectional
morphology. Research on neural cross-lingual transfer shows that
morphological relatedness within a language family directly
facilitates knowledge transfer~\cite{kann2017oneshot}, suggesting
that our results for Polish provide a reasonable lower bound for
the broader Slavic family.

\textbf{Limitations.} Labels are inherited from English annotations without re-annotation in Polish, which precludes measuring cross-cultural label shift; a native-speaker re-annotation study on a subsample would allow this in future work. The pipeline was validated on social media data - Reddit and Twitter discourse styles; domain transfer to news or parliamentary corpora may require re-evaluation, though the diversity of topics covered --- everyday moral discourse, political debate, social movements, and natural disasters --- suggests reasonable robustness across common domains of moral language use. We evaluate only one translation approach (Claude Sonnet); comparing against other MT systems or fine-tuned multilingual models is left for future work. Dialect-heavy content (like AAVE) remains hardest to translate; dialect-aware preprocessing or human post-editing could help here.

\section{Conclusion}
We presented a validated pipeline for extending moral
corpora from English to Polish via LLM translation. Testing across a diverse
range of topics and MFT subcorpora, we find
that translation preserves subtle moral cues well enough for
cross-lingual machine learning to harvest them --- with AUROC gaps typically in the range \textbf{0.01--0.02} under fine-tuning, 
further assessed by DeLong's statistical test.

\textbf{Moral semantics survive machine translation}, opening a practical and
cost-effective path for moral values research in Polish and, by
extension, the broader Slavic family. Beyond the translated corpora themselves, our validation
protocol --- LLM-as-judge auditing, embedding similarity, representational
alignment, and classifier parity --- offers a reusable template for vetting
machine-translated resources in other low-resource NLP settings. Evaluation
of other Slavic languages (e.g., Russian) is planned as follow-up work.

\subsection*{Data and Code Availability.}
Code and supplementary materials for this paper are available at the Open Science Framework: \url{https://osf.io/kp3bj}. 
The translated corpora are available upon reasonable request for academic use.

\begin{acks}
 The author thanks the anonymous reviewers of GoodIT'26 for valuable feedback,
 Adelaide Danilov (University of Luxembourg) for helpful insights on
 translation subtleties in Slavic languages, and Maksymilian Odrzywołek
 along with the University of Warsaw Science Daily team for their
 coverage of this work~\cite{odrzywolek_polscy_2026}. Thanks also to
 journalist Monika Suszek of Polish Radio for covering
 this research in the podcast (Polish Radio Three, "Piknik Naukowy").
\end{acks}

\bibliographystyle{ACM-Reference-Format}
\bibliography{citations}

@article{haidt2004intuitive,
	title = {Intuitive ethics: how innately prepared intuitions generate culturally variable virtues},
	volume = {133},
	issn = {0011-5266, 1548-6192},
	shorttitle = {Intuitive ethics},
	url = {https://direct.mit.edu/daed/article/133/4/55-66/27470},
	doi = {10.1162/0011526042365555},
	language = {en},
	number = {4},
	urldate = {2025-03-13},
	journal = {Daedalus},
	author = {Haidt, Jonathan and Joseph, Craig},
	month = sep,
	year = {2004},
	pages = {55--66},
}

@article{araque_moralstrength_2020,
	title = {{MoralStrength}: {Exploiting} a moral lexicon and embedding similarity for moral foundations prediction},
	volume = {191},
	issn = {09507051},
	shorttitle = {{MoralStrength}},
	url = {https://linkinghub.elsevier.com/retrieve/pii/S095070511930526X},
	doi = {10.1016/j.knosys.2019.105184},
	language = {en},
	urldate = {2026-08-01},
	journal = {Knowledge-Based Systems},
	author = {Araque, Oscar and Gatti, Lorenzo and Kalimeri, Kyriaki},
	month = mar,
	year = {2020},
	pages = {105184},
}

@article{matsuo_development_2019,
	title = {Development and validation of the {Japanese} {Moral} {Foundations} {Dictionary}},
	volume = {14},
	issn = {1932-6203},
	url = {https://dx.plos.org/10.1371/journal.pone.0213343},
	doi = {10.1371/journal.pone.0213343},
	language = {en},
	number = {3},
	urldate = {2026-08-04},
	journal = {PLOS ONE},
	author = {Matsuo, Akiko and Sasahara, Kazutoshi and Taguchi, Yasuhiro and Karasawa, Minoru},
	editor = {Gruebner, Oliver},
	month = mar,
	year = {2019},
	pages = {e0213343},
}

@article{kennedy_moral_2023,
	title = {The (moral) language of hate},
	volume = {2},
	copyright = {https://creativecommons.org/licenses/by/4.0/},
	issn = {2752-6542},
	url = {https://academic.oup.com/pnasnexus/article/doi/10.1093/pnasnexus/pgad210/7220655},
	doi = {10.1093/pnasnexus/pgad210},
	language = {en},
	number = {7},
	urldate = {2026-08-01},
	journal = {PNAS Nexus},
	author = {Kennedy, Brendan and Golazizian, Preni and Trager, Jackson and Atari, Mohammad and Hoover, Joe and Mostafazadeh Davani, Aida and Dehghani, Morteza},
	editor = {Van Bavel, J},
	month = jul,
	year = {2023},
	pages = {pgad210},
}

@inproceedings{feng2020language,
	address = {Dublin, Ireland},
	title = {Language-agnostic {BERT} {Sentence} {Embedding}},
	url = {https://aclanthology.org/2022.acl-long.62},
	doi = {10.18653/v1/2022.acl-long.62},
	language = {en},
	urldate = {2026-07-22},
	booktitle = {Proceedings of the 60th {Annual} {Meeting} of the {Association} for {Computational} {Linguistics} ({Volume} 1: {Long} {Papers})},
	publisher = {Association for Computational Linguistics},
	author = {Feng, Fangxiaoyu and Yang, Yinfei and Cer, Daniel and Arivazhagan, Naveen and Wang, Wei},
	year = {2022},
	pages = {878--891},
}

@inproceedings{kornblith2019similarity,
  title={Similarity of neural network representations revisited},
  author={Kornblith, Simon and Norouzi, Mohammad and Lee, Honglak and Hinton, Geoffrey},
  booktitle={Proceedings of the 36th International Conference on Machine Learning},
  pages={3519--3529},
  year={2019}
}

@inproceedings{devlin2019bert,
	address = {Minneapolis, Minnesota},
	title = {{BERT}: {Pre}-training of {Deep} {Bidirectional} {Transformers} for {Language} {Understanding}},
	url = {http://aclweb.org/anthology/N19-1423},
	doi = {10.18653/v1/N19-1423},
	language = {en},
	urldate = {2026-07-22},
	booktitle = {Proceedings of the 2019 {Conference} of the {North}},
	publisher = {Association for Computational Linguistics},
	author = {Devlin, Jacob and Chang, Ming-Wei and Lee, Kenton and Toutanova, Kristina},
	year = {2019},
	pages = {4171--4186},
}

@inproceedings{conneau2020unsupervised,
  title={Unsupervised cross-lingual representation learning at scale},
  author={Conneau, Alexis and Khandelwal, Kartikay and Goyal, Naman and Chaudhary, Vishrav and Wenzek, Guillaume and Guzm{\'a}n, Francisco and Grave, Edouard and Ott, Myle and Zettlemoyer, Luke and Stoyanov, Veselin},
  booktitle={Proceedings of ACL},
  pages={8440--8451},
  year={2020}
}

@incollection{graham_moral_2013,
	title = {Moral {Foundations} {Theory}: {The} {Pragmatic} {Validity} of {Moral} {Pluralism}},
	volume = {47},
	isbn = {978-0-12-407236-7},
	url = {https://linkinghub.elsevier.com/retrieve/pii/B9780124072367000024},
	doi = {10.1016/B978-0-12-407236-7.00002-4},
	language = {en},
	urldate = {2026-07-22},
	booktitle = {Advances in {Experimental} {Social} {Psychology}},
	publisher = {Elsevier},
	author = {Graham, Jesse and Haidt, Jonathan and Koleva, Sena and Motyl, Matt and Iyer, Ravi and Wojcik, Sean P. and Ditto, Peter H.},
	year = {2013},
	pages = {55--130},
}

@incollection{graham2013moral,
	title = {Moral {Foundations} {Theory}: {The} {Pragmatic} {Validity} of {Moral} {Pluralism}},
	volume = {47},
	isbn = {978-0-12-407236-7},
	url = {https://linkinghub.elsevier.com/retrieve/pii/B9780124072367000024},
	doi = {10.1016/B978-0-12-407236-7.00002-4},
	language = {en},
	urldate = {2026-07-22},
	booktitle = {Advances in {Experimental} {Social} {Psychology}},
	publisher = {Elsevier},
	author = {Graham, Jesse and Haidt, Jonathan and Koleva, Sena and Motyl, Matt and Iyer, Ravi and Wojcik, Sean P. and Ditto, Peter H.},
	year = {2013},
	pages = {55--130},
}

@article{nicholas2023lost,
  title={Lost in Translation: Large Language Models in Non-English Content Analysis},
  author={Nicholas, Gabriel and Bhatia, Aliya},
  journal={arXiv e-prints},
  pages={arXiv--2306},
  year={2023}
}

@inproceedings{plazadelarco2024angry,
	address = {Bangkok, Thailand},
	title = {Angry {Men}, {Sad} {Women}: {Large} {Language} {Models} {Reflect} {Gendered} {Stereotypes} in {Emotion} {Attribution}},
	shorttitle = {Angry {Men}, {Sad} {Women}},
	url = {https://aclanthology.org/2024.acl-long.415},
	doi = {10.18653/v1/2024.acl-long.415},
	language = {en},
	urldate = {2026-07-22},
	booktitle = {Proceedings of the 62nd {Annual} {Meeting} of the {Association} for {Computational} {Linguistics} ({Volume} 1: {Long} {Papers})},
	publisher = {Association for Computational Linguistics},
	author = {Plaza Del Arco, Flor Miriam and Curry, Amanda and Cercas Curry, Alba and Abercrombie, Gavin and Hovy, Dirk},
	year = {2024},
	pages = {7682--7696},
}

@article{graham2009liberals,
  title={Liberals and conservatives rely on different sets of moral foundations},
  author={Graham, Jesse and Haidt, Jonathan and Nosek, Brian A},
  journal={Journal of personality and social psychology},
  volume={96},
  number={5},
  pages={1029--1046},
  year={2009},
  publisher={American Psychological Association},
    doi = {10.1037/a0015141},
}

@article{feinberg2013moral,
	title = {The {Moral} {Roots} of {Environmental} {Attitudes}},
	volume = {24},
	issn = {0956-7976, 1467-9280},
	url = {https://journals.sagepub.com/doi/10.1177/0956797612449177},
	doi = {10.1177/0956797612449177},
	language = {en},
	number = {1},
	urldate = {2026-07-22},
	journal = {Psychological Science},
	author = {Feinberg, Matthew and Willer, Robb},
	month = jan,
	year = {2013},
	pages = {56--62},
}

@article{aksoyWhoseMoralityThey2025,
	title = {Whose morality do they speak? {Unraveling} cultural bias in multilingual language models},
	volume = {12},
	issn = {2949-7191},
	shorttitle = {Whose morality do they speak?},
	url = {https://www.sciencedirect.com/science/article/pii/S2949719125000482},
	doi = {10.1016/j.nlp.2025.100172},
	urldate = {2025-08-26},
	journal = {Natural Language Processing Journal},
	author = {Aksoy, Meltem},
	month = sep,
	year = {2025},
	pages = {100172},
}

@article{grahamMappingMoralDomain2011,
	title = {Mapping the moral domain.},
	volume = {101},
	issn = {1939-1315, 0022-3514},
	url = {https://doi.apa.org/doi/10.1037/a0021847},
	doi = {10.1037/a0021847},
	language = {en},
	number = {2},
	urldate = {2025-03-12},
	journal = {Journal of Personality and Social Psychology},
	author = {Graham, Jesse and Nosek, Brian A. and Haidt, Jonathan and Iyer, Ravi and Koleva, Spassena and Ditto, Peter H.},
	year = {2011},
	pages = {366--385},
}

@inproceedings{conneau_emerging_2020,
	address = {Online},
	title = {Emerging {Cross}-lingual {Structure} in {Pretrained} {Language} {Models}},
	url = {https://aclanthology.org/2020.acl-main.536/},
	doi = {10.18653/v1/2020.acl-main.536},
	booktitle = {Proceedings of the 58th {Annual} {Meeting} of the {Association} for {Computational} {Linguistics}},
	publisher = {Association for Computational Linguistics},
	author = {Conneau, Alexis and Wu, Shijie and Li, Haoran and Zettlemoyer, Luke and Stoyanov, Veselin},
	editor = {Jurafsky, Dan and Chai, Joyce and Schluter, Natalie and Tetreault, Joel},
	month = jul,
	year = {2020},
	pages = {6022--6034},
}

@inproceedings{kumari_quality_2024,
	address = {AU-KBC Research Centre, Chennai, India},
	title = {Quality {Estimation} of {Machine} {Translated} {Texts} based on {Direct} {Evidence} {Approach}},
	url = {https://aclanthology.org/2024.icon-1.16/},
	booktitle = {Proceedings of the 21st {International} {Conference} on {Natural} {Language} {Processing} ({ICON})},
	publisher = {NLP Association of India (NLPAI)},
	author = {Kumari, Vibhuti and Kavi, Narayana Murthy},
	editor = {Lalitha Devi, Sobha and Arora, Karunesh},
	month = dec,
	year = {2024},
	pages = {139--148},
}

@inproceedings{steingrimsson_sentence_2023,
	address = {Singapore},
	title = {A {Sentence} {Alignment} {Approach} to {Document} {Alignment} and {Multi}-faceted {Filtering} for {Curating} {Parallel} {Sentence} {Pairs} from {Web}-crawled {Data}},
	url = {https://aclanthology.org/2023.wmt-1.38},
	doi = {10.18653/v1/2023.wmt-1.38},
	language = {en},
	urldate = {2026-08-02},
	booktitle = {Proceedings of the {Eighth} {Conference} on {Machine} {Translation}},
	publisher = {Association for Computational Linguistics},
	author = {Steingrimsson, Steinthor},
	year = {2023},
	pages = {366--374},
}

@inproceedings{kocmi_large_2023,
	address = {Tampere, Finland},
	title = {Large {Language} {Models} {Are} {State}-of-the-{Art} {Evaluators} of {Translation} {Quality}},
	url = {https://aclanthology.org/2023.eamt-1.19/},
	booktitle = {Proceedings of the 24th {Annual} {Conference} of the {European} {Association} for {Machine} {Translation}},
	publisher = {European Association for Machine Translation},
	author = {Kocmi, Tom and Federmann, Christian},
	editor = {Nurminen, Mary and Brenner, Judith and Koponen, Maarit and Latomaa, Sirkku and Mikhailov, Mikhail and Schierl, Frederike and Ranasinghe, Tharindu and Vanmassenhove, Eva and Vidal, Sergi Alvarez and Aranberri, Nora and Nunziatini, Mara and Escartín, Carla Parra and Forcada, Mikel and Popovic, Maja and Scarton, Carolina and Moniz, Helena},
	month = jun,
	year = {2023},
	pages = {193--203},
}

@inproceedings{alain_understanding_2017,
	title = {Understanding intermediate layers using linear classifier probes},
	url = {https://openreview.net/forum?id=HJ4-rAVtl},
	booktitle = {5th {International} {Conference} on {Learning} {Representations}, {ICLR} 2017, {Toulon}, {France}, {April} 24-26, 2017, {Workshop} {Track} {Proceedings}},
	publisher = {OpenReview.net},
	author = {Alain, Guillaume and Bengio, Yoshua},
	year = {2017},
}

@inproceedings{he_debertav3_2023,
	title = {{DeBerTaV3}: {Improving} {DeBerTa} using {ELECTRA}-{Style} {Pre}-{Training} with {Gradient}-{Disentangled} {Embedding} {Sharing}},
	url = {https://openreview.net/forum?id=sE7-XhLxHA},
	booktitle = {International {Conference} on {Learning} {Representations}},
	author = {He, Pengcheng and Gao, Jianfeng and Chen, Weizhu},
	year = {2023},
}

@article{nguyen2022mapping,
	title = {Mapping {Topics} in 100,000 {Real}-{Life} {Moral} {Dilemmas}},
	volume = {16},
	issn = {2334-0770, 2162-3449},
	url = {https://ojs.aaai.org/index.php/ICWSM/article/view/19327},
	doi = {10.1609/icwsm.v16i1.19327},
	urldate = {2026-07-22},
	journal = {Proceedings of the International AAAI Conference on Web and Social Media},
	author = {Nguyen, Tuan Dung and Lyall, Georgiana and Tran, Alasdair and Shin, Minjeong and Carroll, Nicholas George and Klein, Colin and Xie, Lexing},
	month = may,
	year = {2022},
	pages = {699--710},
}

@book{haidt2012righteous,
  title = {The {{Righteous Mind}}: {{Why Good People Are Divided}} by {{Politics}} and {{Religion}}},
  shorttitle = {The {{Righteous Mind}}},
  author = {Haidt, Jonathan},
  year = {2012},
  month = mar,
  publisher = {Knopf Doubleday Publishing Group},
  googlebooks = {ItuzJhbcpMIC},
  isbn = {978-0-307-90703-5},
  langid = {english}
}

@inproceedings{kann2017oneshot,
    title = "One-Shot Neural Cross-Lingual Transfer for Paradigm Completion",
    author = {Kann, Katharina  and
      Cotterell, Ryan  and
      Sch{\"u}tze, Hinrich},
    editor = "Barzilay, Regina  and
      Kan, Min-Yen",
    booktitle = "Proceedings of the 55th Annual Meeting of the Association for Computational Linguistics (Volume 1: Long Papers)",
    month = jul,
    year = "2017",
    address = "Vancouver, Canada",
    publisher = "Association for Computational Linguistics",
    url = "https://aclanthology.org/P17-1182/",
    doi = "10.18653/v1/P17-1182",
    pages = "1993--2003"
}

@article{nguyenMeasuringMoralDimensions2024a,
  title = {Measuring {{Moral Dimensions}} in {{Social Media}} with {{Mformer}}},
  author = {Nguyen, Tuan Dung and Chen, Ziyu and Carroll, Nicholas George and Tran, Alasdair and Klein, Colin and Xie, Lexing},
  year = {2024},
  month = may,
  journal = {Proceedings of the International AAAI Conference on Web and Social Media},
  volume = {18},
  pages = {1134--1147},
  issn = {2334-0770, 2162-3449},
  doi = {10.1609/icwsm.v18i1.31378},
  url = {https://ojs.aaai.org/index.php/ICWSM/article/view/31378},
  urldate = {2025-05-29}
}

@inproceedings{preniqiMoralBERTFineTunedLanguage2024,
  title = {{{MoralBERT}}: {{A Fine-Tuned Language Model}} for {{Capturing Moral Values}} in {{Social Discussions}}},
  shorttitle = {{{MoralBERT}}},
  booktitle = {Proceedings of the 2024 {{International Conference}} on {{Information Technology}} for {{Social Good}}},
  author = {Preniqi, Vjosa and Ghinassi, Iacopo and Ive, Julia and Saitis, Charalampos and Kalimeri, Kyriaki},
  year = {2024},
  month = sep,
  pages = {433--442},
  publisher = {ACM},
  address = {Bremen Germany},
  doi = {10.1145/3677525.3678694},
  url = {https://dl.acm.org/doi/10.1145/3677525.3678694},
  urldate = {2025-05-29},
  isbn = {9798400710940},
  langid = {english}
}

@inproceedings{zangari-etal-2025-me2,
  title = {{{ME2-BERT}}: {{Are}} Events and Emotions What You Need for Moral Foundation Prediction?},
  booktitle = {Proceedings of the 31st International Conference on Computational Linguistics},
  author = {Zangari, Lorenzo and Greco, Candida M. and Picca, Davide and Tagarelli, Andrea},
  editor = {Rambow, Owen and Wanner, Leo and Apidianaki, Marianna and {Al-Khalifa}, Hend and Eugenio, Barbara Di and Schockaert, Steven},
  year = {2025},
  month = jan,
  pages = {9516--9532},
  publisher = {Association for Computational Linguistics},
  address = {Abu Dhabi, UAE},
  url = {https://aclanthology.org/2025.coling-main.638/}
}

@inproceedings{skorski-landowska-2025-beyond,
    title = "Beyond Human Judgment: A {B}ayesian Evaluation of {LLM}s' Moral Values Understanding",
    author = "Skorski, Maciej  and
      Landowska, Alina",
    booktitle = "Proceedings of the 2nd Workshop on Uncertainty-Aware NLP (UncertaiNLP 2025)",
    month = nov,
    year = "2025",
    address = "Suzhou, China",
    publisher = "Association for Computational Linguistics",
    url = "https://aclanthology.org/2025.uncertainlp-main.3/",
    doi = "10.18653/v1/2025.uncertainlp-main.3",
    pages = "17--26",
    ISBN = "979-8-89176-349-4"
}

@article{hooverMoralFoundationsTwitter2020,
  title = {Moral {{Foundations Twitter Corpus}}: {{A Collection}} of 35k {{Tweets Annotated}} for {{Moral Sentiment}}},
  shorttitle = {Moral {{Foundations Twitter Corpus}}},
  author = {Hoover, Joe and {Portillo-Wightman}, Gwenyth and Yeh, Leigh and Havaldar, Shreya and Davani, Aida Mostafazadeh and Lin, Ying and Kennedy, Brendan and Atari, Mohammad and Kamel, Zahra and Mendlen, Madelyn and Moreno, Gabriela and Park, Christina and Chang, Tingyee E. and Chin, Jenna and Leong, Christian and Leung, Jun Yen and Mirinjian, Arineh and Dehghani, Morteza},
  year = {2020},
  month = nov,
  journal = {Social Psychological and Personality Science},
  volume = {11},
  number = {8},
  pages = {1057--1071},
  issn = {1948-5506, 1948-5514},
  doi = {10.1177/1948550619876629},
  url = {https://journals.sagepub.com/doi/10.1177/1948550619876629},
  urldate = {2025-08-11},
  langid = {english}
}

@article{haidtWhenMoralityOpposes2007,
	title = {When {Morality} {Opposes} {Justice}: {Conservatives} {Have} {Moral} {Intuitions} that {Liberals} may not {Recognize}},
	volume = {20},
	copyright = {http://www.springer.com/tdm},
	issn = {0885-7466, 1573-6725},
	shorttitle = {When {Morality} {Opposes} {Justice}},
	url = {http://link.springer.com/10.1007/s11211-007-0034-z},
	doi = {10.1007/s11211-007-0034-z},
	language = {en},
	number = {1},
	urldate = {2025-03-13},
	journal = {Social Justice Research},
	author = {Haidt, Jonathan and Graham, Jesse},
	month = jun,
	year = {2007},
	pages = {98--116},
}

@article{atariMoralityWEIRDHow2023,
	title = {Morality beyond the {WEIRD}: {How} the nomological network of morality varies across cultures.},
	volume = {125},
	copyright = {http://www.apa.org/pubs/journals/resources/open-access.aspx},
	issn = {1939-1315, 0022-3514},
	shorttitle = {Morality beyond the {WEIRD}},
	url = {https://doi.apa.org/doi/10.1037/pspp0000470},
	doi = {10.1037/pspp0000470},
	language = {en},
	number = {5},
	urldate = {2025-12-21},
	journal = {Journal of Personality and Social Psychology},
	author = {Atari, Mohammad and Haidt, Jonathan and Graham, Jesse and Koleva, Sena and Stevens, Sean T. and Dehghani, Morteza},
	month = nov,
	year = {2023},
	note = {https://osf.io/srtxn},
	pages = {1157--1188},
}

@article{graham_beyond_2010,
	title = {Beyond {Beliefs}: {Religions} {Bind} {Individuals} {Into} {Moral} {Communities}},
	volume = {14},
	copyright = {https://journals.sagepub.com/page/policies/text-and-data-mining-license},
	issn = {1088-8683, 1532-7957},
	shorttitle = {Beyond {Beliefs}},
	url = {https://journals.sagepub.com/doi/10.1177/1088868309353415},
	doi = {10.1177/1088868309353415},
	language = {en},
	number = {1},
	urldate = {2026-08-01},
	journal = {Personality and Social Psychology Review},
	author = {Graham, Jesse and Haidt, Jonathan},
	month = feb,
	year = {2010},
	pages = {140--150},
}

@article{graham_liberals_2009,
	title = {Liberals and conservatives rely on different sets of moral foundations.},
	volume = {96},
	issn = {1939-1315, 0022-3514},
	url = {https://doi.apa.org/doi/10.1037/a0015141},
	doi = {10.1037/a0015141},
	language = {en},
	number = {5},
	urldate = {2026-08-01},
	journal = {Journal of Personality and Social Psychology},
	author = {Graham, Jesse and Haidt, Jonathan and Nosek, Brian A.},
	month = may,
	year = {2009},
	pages = {1029--1046},
}

@article{kivikangas_moral_2021,
	title = {Moral foundations and political orientation: {Systematic} review and meta-analysis.},
	volume = {147},
	issn = {1939-1455, 0033-2909},
	shorttitle = {Moral foundations and political orientation},
	url = {https://doi.apa.org/doi/10.1037/bul0000308},
	doi = {10.1037/bul0000308},
	language = {en},
	number = {1},
	urldate = {2026-08-01},
	journal = {Psychological Bulletin},
	author = {Kivikangas, J. Matias and Fernández-Castilla, Belén and Järvelä, Simo and Ravaja, Niklas and Lönnqvist, Jan-Erik},
	month = jan,
	year = {2021},
	pages = {55--94},
}

@article{skorskiMoralGapLarge2025,
  title = {The {{Moral Gap}} of {{Large Language Models}}},
  author = {Skorski, Maciej and Landowska, Alina},
  year = {2025},
  eprint = {2507.18523},
  primaryclass = {cs},
  doi = {10.13140/RG.2.2.26221.70880},
  url = {http://arxiv.org/abs/2507.18523},
  urldate = {2025-08-19},
  archiveprefix = {arXiv}
}

@inproceedings{royAnalysisNuancedStances2021,
  title = {Analysis of {{Nuanced Stances}} and {{Sentiment Towards Entities}} of {{US Politicians}} through the {{Lens}} of {{Moral Foundation Theory}}},
  booktitle = {Proceedings of the {{Ninth International Workshop}} on {{Natural Language Processing}} for {{Social Media}}},
  author = {Roy, Shamik and Goldwasser, Dan},
  year = {2021},
  pages = {1--13},
  publisher = {Association for Computational Linguistics},
  address = {Online},
  doi = {10.18653/v1/2021.socialnlp-1.1},
  url = {https://www.aclweb.org/anthology/2021.socialnlp-1.1},
  urldate = {2025-08-16},
  langid = {english}
}

@article{hoppExtendedMoralFoundations2021,
  title = {The Extended {{Moral Foundations Dictionary}} ({{eMFD}}): {{Development}} and Applications of a Crowd-Sourced Approach to Extracting Moral Intuitions from Text},
  shorttitle = {The Extended {{Moral Foundations Dictionary}} ({{eMFD}})},
  author = {Hopp, Frederic R. and Fisher, Jacob T. and Cornell, Devin and Huskey, Richard and Weber, Ren{\'e}},
  year = {2021},
  month = feb,
  journal = {Behavior Research Methods},
  volume = {53},
  number = {1},
  pages = {232--246},
  issn = {1554-3528},
  doi = {10.3758/s13428-020-01433-0},
  url = {https://link.springer.com/10.3758/s13428-020-01433-0},
  urldate = {2025-05-29},
  langid = {english}
}

@misc{tragerMoralFoundationsReddit2022,
  title = {The {{Moral Foundations Reddit Corpus}}},
  author = {Trager, Jackson and Ziabari, Alireza S. and Davani, Aida Mostafazadeh and Golazizian, Preni and {Karimi-Malekabadi}, Farzan and Omrani, Ali and Li, Zhihe and Kennedy, Brendan and Reimer, Nils Karl and Reyes, Melissa and Cheng, Kelsey and Wei, Mellow and Merrifield, Christina and Khosravi, Arta and Alvarez, Evans and Dehghani, Morteza},
  year = {2022},
  publisher = {arXiv},
  doi = {10.48550/ARXIV.2208.05545},
  url = {https://arxiv.org/abs/2208.05545},
  urldate = {2025-05-29},
  copyright = {Creative Commons Attribution 4.0 International}
}

@article{tirkkonenCondit2002,
	title = {Translationese – a myth or an empirical fact?: {A} study into the linguistic identifiability of translated language},
	volume = {14},
	issn = {0924-1884, 1569-9986},
	shorttitle = {Translationese – a myth or an empirical fact?},
	url = {https://benjamins.com/online/target/articles/target.14.2.02tir},
	doi = {10.1075/target.14.2.02tir},
	language = {en},
	number = {2},
	urldate = {2026-07-20},
	journal = {Target. International Journal of Translation Studies},
	author = {Tirkkonen-Condit, Sonja},
	month = dec,
	year = {2002},
	pages = {207--220},
}

@inproceedings{wein-schneider-2024-lost,
	address = {St. Julian’s, Malta},
	title = {Lost in {Translationese}? {Reducing} {Translation} {Effect} {Using} {Abstract} {Meaning} {Representation}},
	shorttitle = {Lost in {Translationese}?},
	url = {https://aclanthology.org/2024.eacl-long.45},
	doi = {10.18653/v1/2024.eacl-long.45},
	language = {en},
	urldate = {2026-07-20},
	booktitle = {Proceedings of the 18th {Conference} of the {European} {Chapter} of the {Association} for {Computational} {Linguistics} ({Volume} 1: {Long} {Papers})},
	publisher = {Association for Computational Linguistics},
	author = {Wein, Shira and Schneider, Nathan},
	year = {2024},
	pages = {753--765},
}

@misc{odrzywolek_polscy_2026,
	title = {Polscy naukowcy uczę sztuczną inteligencję moralnosci},
	url = {https://serwisnaukowy.uw.edu.pl/polscy-naukowcy-ucza-sztuczna-inteligencje-moralnosci/},
	language = {polish},
	journal = {UW Science Daily},
	author = {Odrzywołek, Maksymilian},
	month = jun,
	year = {2026},
}

@article{delong_comparing_1988,
	title = {Comparing the {Areas} under {Two} or {More} {Correlated} {Receiver} {Operating} {Characteristic} {Curves}: {A} {Nonparametric} {Approach}},
	volume = {44},
	issn = {0006341X},
	shorttitle = {Comparing the {Areas} under {Two} or {More} {Correlated} {Receiver} {Operating} {Characteristic} {Curves}},
	url = {https://www.jstor.org/stable/2531595?origin=crossref},
	doi = {10.2307/2531595},
	number = {3},
	urldate = {2026-07-22},
	journal = {Biometrics},
	author = {DeLong, Elizabeth R. and DeLong, David M. and Clarke-Pearson, Daniel L.},
	month = sep,
	year = {1988},
	pages = {837},
}

@article{sun_fast_2014,
	title = {Fast {Implementation} of {DeLong}’s {Algorithm} for {Comparing} the {Areas} {Under} {Correlated} {Receiver} {Operating} {Characteristic} {Curves}},
	volume = {21},
	copyright = {https://ieeexplore.ieee.org/Xplorehelp/downloads/license-information/IEEE.html},
	issn = {1070-9908, 1558-2361},
	url = {http://ieeexplore.ieee.org/document/6851192/},
	doi = {10.1109/LSP.2014.2337313},
	number = {11},
	urldate = {2026-07-22},
	journal = {IEEE Signal Processing Letters},
	author = {Sun, Xu and Xu, Weichao},
	month = nov,
	year = {2014},
	pages = {1389--1393},
}

@article{davidson2017automated,
	title = {Automated {Hate} {Speech} {Detection} and the {Problem} of {Offensive} {Language}},
	volume = {11},
	issn = {2334-0770, 2162-3449},
	url = {https://ojs.aaai.org/index.php/ICWSM/article/view/14955},
	doi = {10.1609/icwsm.v11i1.14955},
	number = {1},
	urldate = {2026-07-22},
	journal = {Proceedings of the International AAAI Conference on Web and Social Media},
	author = {Davidson, Thomas and Warmsley, Dana and Macy, Michael and Weber, Ingmar},
	month = may,
	year = {2017},
	pages = {512--515},
}

\end{document}